# Exploring Effective Strategies for Building a User-Configured GPT for Coding Classroom Dialogues

Luwei Bai, Dongkeun Han, Sara Hennessy

*Faculty of Education, University of Cambridge, UK*

## Abstract

This study investigated effective strategies for developing a custom GPT to code classroom dialogue. While classroom dialogue is widely recognised as a crucial element of education, its analysis remains challenging due to the need for a nuanced understanding of dialogic functions and the labour-intensive nature of manual transcript coding. Recent advancements in large language models (LLMs) offer promising avenues for automating this process. However, existing studies predominantly focus on training large-scale models or evaluating pre-trained models with fixed codebooks, the outcomes of which are often not applicable, or the methods are not replicable for dialogue researchers working with small datasets or employing customised coding schemes. Using MyGPT – a GPT-4-based customised system configured for dialogue analysis – as a case, this study evaluates its baseline performance in coding classroom dialogue with a human codebook and examines how performance varies with different example inputs under a controlled variable design. Through a design-based research approach, this study explores a set of practical strategies – based upon MyGPT's unique features – for configuring an effective tool with limited data. The findings suggest that, despite a few limitations, a custom GPT developed using these specific strategies can serve as a useful coding assistant by generating coding suggestions.

# 1. Introduction

## 1.1 Classroom Dialogue: A Vibrant and Evolving Field of Study

Classroom dialogue—underpinned by essential elements such as idea exchange and evaluation, collaborative knowledge building, logical reasoning, evidence-based justification, and conceptual elaboration (Nystrand, Wu, Gamoran, Zeiser, & Long 2003; Michaels & O'Connor, 2015)—is widely acknowledged as a vital tool for learning (Howe & Abedin, 2013). A growing body of empirical research (e.g., Alexander, 2018; Howe, Hennessy, Mercer, Vrikki, & Wheatley, 2019; Kuhn, 2015) has demonstrated that dialogic pedagogies can substantially enhance learning outcomes; these approaches not only lead to improved academic attainment across subjects (Howe et al., 2019), but also facilitate higher-order thinking (Kuhn, 2015) and deep conceptual understanding (Resnick, Michaels, and O'Connor, 2010). Moreover, dialogue-rich classrooms foster stronger communication skills and collaborative engagement among learners (O'Connor & Michaels, 2007). Reflecting these broad benefits, classroom dialogue has emerged globally as a vibrant area of study, encompassing a range of research traditions, focal interests, and methodological approaches (Kershner et al., 2020).

## 1.2 Classroom Dialogue Analysis through Coding Schemes

As interest in dialogue research has grown, various efforts have been made to establish systematic methods for its analysis. Chief among them is the development of *coding* frameworks – which offer a *structured*, *theory-informed* approach for assessing both the quantity and quality of dialogue in educational settings (Hennessy, Howe, Mercer, & Vrikki, 2020). One prominent series includes the Scheme for Educational Dialogue Analysis (SEDA) (Hennessy et al., 2016), the Toolkit for Systematic Educational Dialogue Analysis (T-SEDA) (T-SEDA Collective, 2023), and the Cambridge Dialogue Analysis Scheme (CDAS) (Vrikki, Wheatley, Howe, & Mercer, 2019). Based upon decades of discourse research, these frameworks categorise specific features of dialogic interaction (e.g., types of invitations, uptake of ideas). When applied to lesson transcripts, such schemes can allow researchers and educators to quantify the dialogicality of classroom exchanges, moving beyond intuition to *concrete* evidence of interaction patterns.

These dialogue analysis schemes typically rely on *deductive* coding (Linneberg & Korsgaard, 2019), with categories derived from established theoretical frameworks and empirical findings. For instance, SEDA is grounded in a socio-cultural paradigm and draws upon concepts from discourse theory (e.g., Hymes' ethnography of communication) to define key dialogic moves (Hennessy et al., 2016). These moves comprise inviting students to build on prior contributions, challenging ideas, and making connections with previous knowledge. By coding each conversational turn according to these *predefined* categories, users can evaluate how dialogic classroom exchanges are – identifying patterns in the frequency and type of dialogic moves and comparing them across lessons or interventions. For researchers, such deductive coding based on established schemes can provide a shared analytical language, support the replication of studies, and contribute to the accumulation of empirical evidence (Howe & Abedin, 2013). For educators, applying these frameworks can disclose habitual interaction patterns and inform efforts to improve specific dimensions of their classroom dialogue (Hennessy, Kershner, & Ahmed, 2021).

### 1.3 Constraints of Coding-Based Dialogue Analysis

Despite the aforementioned benefits, manual coding of classroom dialogue is "an immensely complex and cognitively demanding activity that has taxed researchers over decades." (Hennessy et al., 2020, p. 1). While recent advances in transcription software have reduced the time required to document recorded dialogues, the coding phase continues to present major difficulties due to its *labour-intensive* and *time-consuming* nature (Long, Luo, & Zhang, 2024). These challenges are not limited to the coding phase itself; prior to applying coding schemes with consistency and accuracy, human coders must receive thorough training.

What should not be overlooked, furthermore, is the possibility that human coders' misinterpretation and potential biases might shape their coding decisions – especially when dealing with abstract categories and ambiguous cases. Such inconsistencies can not only affect the reliability of the coding-based analysis of classroom dialogue but also raise concerns about the validity of the conclusions drawn from it.

### 1.4 The Emergence of AI Tools: A Game-Changer in Dialogue Analysis?

Recent developments in artificial intelligence (AI) provide promising opportunities to address the challenges associated with qualitative data analysis. Particularly, LLMs and related deep-learning techniques have exhibited considerable capacity in understanding and analysing human language, kindling growing interest in their application to deductive qualitative coding. (e.g., Na & Feng, 2025; Zhang et al., 2025).

Building on this momentum, a few dialogue researchers have explored whether

– and how – LLMs or other deep learning approaches can automate classroom dialogue coding, functioning as AI coders on par with highly trained human annotators. For instance, Long et al (2024), examined a customised GPT-4 that coded dialogues from 6 mathematics and 6 language lessons in Chinese middle schools, using a revised version of the CDAS scheme. The AI-generated coding outputs were compared with expert humans' manual annotations of the same data. The results indicate that the GPT-4-based approach achieved high alignment with human coders while significantly reducing the time needed for analysis. Similarly, Song et al. (2021) developed and assessed an artificial neural network-based classifier – trained on utterances manually labelled by human coders and tailored specially for a predefined set of seven semantic categories (e.g., agreement, uptake, speculation). This study drew upon 155 recorded lessons (mathematics, science, and literacy) from 74 primary and 81 secondary schools in China. This classifier's overall performance was found to be comparable to that of the human coders. What should not be overlooked, nonetheless, is the agreement for the categories of querying and uptake remained relatively low (precision 0.438 and 0.584, respectively) despite the central role these talk moves play in classroom dialogues.

These studies point to the potential of AI technologies to enhance the efficiency and consistency of dialogue coding in education research. It is also to be noted, however, that not only is inter-coder reliability variable, but their practical *replicability* remains *limited*, and these studies focused more on presenting the effectiveness of the ready-made AI dialogue coders rather than exploring potential strategies.  Long et al. (2024) offer only a general description of the specialised and prompt-engineering – without providing details about the underlying technical procedures – required to reproduce their AI coder. Although Song et al.'s (2021) methodology is more transparent, it is built on a complex process involving pre-trained transformer-based embeddings (BERT) and a substantial dataset of 155 recorded lessons – resources that are rarely available to many researchers, even less so to educators. Therefore, the gap persists between experimental demonstrations of AI-based dialogue coding and its *realistic* implementation in everyday educational settings – a concern made sharper by the fact that effective dialogue analysis usually requires coding schemes *tailored* to *local* curricula and contexts.

The present study aimed to close this gap by developing and openly documenting practical, specific strategies that allow both researchers and teachers to build *lightweight* AI assistants for dialogue coding – without specialised infrastructure. Drawing upon the CDAS framework and a *modest* set of human-coded scripts, we tested several approaches to constructing a custom GPT – a low-cost AI system that can be easily configured by non-specialists to perform specific tasks. By disclosing these comprehensive, viable strategies, we intended to lower the entry barrier for dialogue researchers seeking to develop AI coding assistants tailored to their specific needs as well as to empower educators to generate context-sensitive diagnostics of their own dialogic practices.

## 2. Background and Relevant Work

### 2.1 Large Language Models

LLMs are a type of artificial intelligence trained on vast amounts of text data and can perform various natural language processing (NLP) tasks like text generation, translation, and summarisation. Deep learning architectures used in LLMs allow them to learn context and meaning from large datasets (Hadi et al., 2023). As a result, they demonstrate strong capability in understanding and generating human language

(Naveed et al., 2023).

The emergence of LLMs signals a transformative shift in qualitative research. These models offer novel avenues for language analysis, promising to streamline the qualitative coding process and mitigate the challenges associated with handling extensive datasets. The potential for LLMs in enhancing qualitative coding has been demonstrated in many recent studies (e.g., Gamieldien et al., 2023).

## 2.2 GPT and Custom GPT

GPTs are developed by OpenAI and use deep learning techniques to understand and generate human-like text. These models are based on the Transformer architecture introduced by Vaswani et al. (2017), enabling efficient processing of sequential data through self-attention mechanisms. They are pretrained using a large volume of publicly available text and subsequently adapted to specific tasks through fine-tuning or in-context learning (Brown et al., 2020). Beginning with GPT-4, newer GPT models have demonstrated significant advances in language understanding and generation, and demonstrate strong performance across a wide range of natural language processing (NLP) benchmarks, including summarisation, translation, question answering, and reasoning (OpenAI, 2023a).

GPT models were initially designed as general-purpose language models trained based on next-token prediction on massive text corpora. However, recent developments have expanded their capabilities to support agentic behaviours, enabling them to act as autonomous tools that can plan, reason, and interact dynamically with their environments or users. This transformation is facilitated through methods such as Reinforcement Learning with Human Feedback (RLHF), tool use, and multi-step reasoning, which allow models to move beyond passive text generation toward interactive decision-making and problem-solving (OpenAI, 2023a).

The lightweight customised GPTs offered by OpenAI are better understood as configurable conversational assistants than as fully autonomous agents. Although they may exhibit limited tool-using or goal-directed behaviours depending on their configurations, they do not necessarily operate within the kind of persistent action-feedback loop typically associated with agent training. These customised GPTs are also subject to token limits depending on the underlying model (OpenAI, 2023b).

Since the emergence of GPT-3, studies have explored the application of GPT models in qualitative analysis and thematic coding. For instance, Xiao et al. (2023) combined GPT-3 with expert-drafted codebooks and achieved fair to substantial agreement with expert-coded results through a specific training approach. Beyond qualitative analysis at the general text level, GPT models also show strong potential for coding classroom dialogue, which is often analysed in relation to the speaker's intention. In Long et al.'s (2024) study, a customised GPT effectively reduced the time required for classroom dialogue coding while achieving a high level of agreement with human coders.

## 2.3 Cognitive Load in Training Pipeline

Coding dialogue is a complex task for both human coders and customised GPT systems (Long et al., 2024). Prior work suggests that LLM performance can degrade when reasoning tasks impose multiple interacting constraints, and some researchers have analysed this through a cognitive-load-inspired lens (Yue et al., 2023). This tendency was also reflected in our initial exploration: when multiple rules were combined, the customised GPT did not always follow all instructions consistently. In addition, LLMs may exhibit conformity effects, becoming more likely to align with external suggestions when uncertain about their own judgments (Zhu et al., 2025).

Researchers have therefore explored methods for reducing reasoning burden and improving instruction efficiency. For example, Selective Propositional Reasoning (SPR) was proposed to streamline reasoning steps and improve performance (Yue et al., 2023). Without changing model architecture, prompt-optimisation methods such as instruction editing with GrIPS and discrete prompt optimisation with RLPrompt can also improve task performance by producing more effective prompts (Prasad et al., 2023; Deng et al., 2022).

### 2.4 Decision Trees for Alleviating Cognitive Load.

Research has demonstrated that the use of decision trees can help reduce cognitive load by structuring complex judgments into smaller, sequential decisions. Rather than requiring a decision-maker to evaluate all variables simultaneously, a tree-based structure presents a limited set of options at each step, which may reduce working-memory demands. This logic aligns with the chunking principle in cognitive psychology, whereby information is organised into manageable units (Miller, 1956), and with cognitive load theory, which emphasises the limits of working memory during problem solving (Sweller, 1988). In LLM applications, decision-tree-like logic may be incorporated through prompt structure or reasoning frameworks. However, ReAct is better understood as a sequential reasoning-and-acting framework than as a tree structure (Yao et al., 2023). If a branching analogy is intended, Tree of Thoughts provides a closer conceptual match because it explicitly explores multiple reasoning paths before selecting the next step (Yao et al., 2023; Zhou et al., 2024).

## 2.5 Structured Segments and Anchor Examples to Improve AI's Performance

Instead of embedding all instructions, definitions, and examples into one long prompt, it may be more effective to separate them into structured segments, such as a definition block, an instruction block, and an examples block. This idea aligns with work on decomposed or modular prompting, which shows that breaking complex tasks into simpler prompt components can improve performance (Khot et al., 2022). In addition, models can sometimes benefit from being prompted to critique and revise their own outputs, as shown in work on self-refinement and self-feedback (Madaan et al., 2023). Relatedly, self-consistency prompting improves reasoning by sampling multiple solution paths and selecting the most consistent answer (Wang et al., 2022). Providing representative examples for each code, including clear positives and contrastive cases, may also help the model distinguish between similar categories. This is consistent with findings that demonstrations help in-context learning by conveying label space, input distribution, and output format (Min et al., 2022).

## 2.6 Research Questions

Summing up the literature review, we can surmise that:

1) Training an LLM needs a large volume of data; even a small language model needs a big volume of data, which is often impractical for many researchers;

2) Using a custom GPT directly for complex tasks like dialogic coding may face inaccuracy caused by cognitive load;

3) Decision trees and modular prompts may be potential strategies to improve the AI dialogue coder's performance on dialogic coding.

In this study, we aimed to explore effective strategies for developing a customised GPT

to support the coding of classroom dialogue. By tuning a MyGPT, which operates without API integration—we introduce approaches tailored to the needs of qualitative researchers working with small datasets. These strategies include optimised instruction design within the model's configurations and informed sample selection. Accordingly, this study is guided by the following three research questions.

RQ1) How does a MyGPT with basic configurations (instructions) perform in analysing classroom dialogue using CDAS for human coders?

RQ2) Does data size matter in the custom GPT setting? How does its performance change with different small sizes of examples?

RQ3) What are the potential strategies for building an effective custom GPT for coding dialogues?

## 3. Method

### 3.1 General process

For RQ1 and RQ2, a controlled variable approach (Bernerth & Aguinis, 2016) was employed to evaluate the AI coder's performance across different example sample sizes. We used a validated coding scheme (see the description of CDAS in 3.2), identical instruction configurations, consistent testing materials with the same human-coded results, and ensured stable sample quality drawn from the same datasets. The only variable manipulated was the number of examples provided. After testing the baseline AI coder (configured with only basic code definitions and the same training materials used for human coders, but without specific prompt configurations or example inputs), we evaluated four experimental conditions, adopting examples from 30 lessons:

1. One example per code category.
2. Ten examples per code type (120 in total), presented in separate turns.
3. A total of 120 examples presented in natural dialogue flows, covering all code types, though not evenly distributed due to their natural occurrence.
4. A total of 500 examples in natural dialogue flows.

The coding outputs generated by a custom GPT under each condition were then compared to the results produced by human coders.

For RQ3, we adopted a design-based approach (Anderson & Shattuck, 2012), which involved identifying potential strategies through a literature review and exploring with GPT, establishing a preliminary framework for implementation, and testing the setup through iterative cycles. The results and reflections from each cycle were then used to inform ongoing refinements and improvements.

The initial design conjectures suggest the following considerations, summing from the literature review:

- Due to token limitations, providing a large number of examples does not necessarily enhance the AI coder's performance. Therefore, careful selection of examples is essential.

- The configuration of instructions plays a more critical role in performance than the quantity of examples provided.

- Reducing cognitive load—particularly in tasks involving complex judgment and reasoning—is crucial for improving performance. A decision-tree-based approach may offer an effective solution.

- Although a MyGPT does not support direct code execution, GPT-4 is capable of understanding and emulating logical structures commonly used in programming, such as if...else statements. Embedding such conditional logic within prompts can clarify classification criteria, support structured decision-making, and reduce ambiguity—potentially enhancing the model's performance in rule-based coding tasks.

- Iterative feedback is important for the model's performance.

- Modularising the instructions and prompts are necessary in enhancing stability.

### 3.2 Coding Framework: The CDAS Scheme

This study adopted the Cambridge Dialogue Analysis scheme (CDAS) as its analytic foundation and coupled it with a Custom GPT that applies the scheme at scale. CDAS is one of the most recent condensed versions of SEDA (the Scheme for Educational Dialogue Analysis). Hennessy et al., (2016) initially proposed the full SEDA framework (comprising 33 codes) and highlighted that it could be reduced to fewer distinct codes without compromising theoretical coverage. Building on this, Howe, Hennessy, Mercer & Vrikki (2019) further refined the phrasing of several categories and consolidated the framework into 13 categories, as outlined in Table 1, first presented by Vrikki et al. (2019).

**Table 1**

*The Cambridge Dialogue Analysis Scheme (CDAS) Code Descriptions*

| Codes | Descriptions |
|---|---|
| Elaboration Invitation (ELI) | Invites others to build on, evaluate, or clarify prior contributions. Excludes unseen work or procedural follow-ups. |
| Elaboration (EL) | Builds on or adds new ideas/perspectives to earlier contributions. Includes brief but meaningful elaborations or related ideas. |
| Reasoning Invitation (IRE) | Asks for explanation, justification, speculation, or prediction (e.g. “Why?”, “What if...?”). Excludes simple answer requests. |
| Reasoning (RE) | Provides reasons, explanations, or evidence for a view. Includes analogies, distinctions, and justified speculations. |
| Co-ordination Invitation (CI) | Invites comparison, synthesis, or resolution of two or more ideas. |
| Simple Co-ordination (SC) | Summarises or compares ideas (own or others’) without giving reasons. |
| Reasoned Co- | Compares or integrates ideas with justification or evidence. |

| ordination (RC) | Includes counter-arguments and reasoned agreement. |
|---|---|
| Agreement (A) | Explicit agreement or acceptance (e.g. “Yes”, “I agree”). Includes paraphrasing or repetition to signal agreement. |
| Querying (Q) | Challenges or disagrees with a statement. Includes verbal disagreement, sarcasm, or questioning. |
| Reference Back (RB) | Refers to prior class knowledge, shared experiences, or earlier activities. |
| Reference to Wider Context (RW) | Links current learning to broader contexts (e.g. real-world examples, outside expertise). |
| Other Invitation (OI) | All other verbal invitations (e.g. ideas, opinions, closed/open questions, calculations). Excludes non-verbal prompts. |
| Uncoded | When none of the above codes apply |

CDAS was selected for two main reasons. First, it is a well-established framework applied successfully in two large-scale studies (i.e., Howe et al., 2019; Vrikki et al., 2019). Howe et al (2019), for example, employed CDAS and confirmed the scheme’s validity across math, science, and English lessons conducted in primary schools in England, based on a sample of 144 lessons with children aged 10-11. The codes were categorised by the dialogic function of the spoken turns rather than the content meaning. Moreover, its revised versions have informed recent AI-driven research on automatic coding of classroom dialogue (Long et al, 2024; Song et al, 2021). Second, we had access to classroom dialogue data from Howe et al. (2019)’s research, which had been manually annotated by trained human coders using CDAS (and inter-coder reliability measured before applying the scheme to the whole dataset). These annotations served as the basis for developing and assessing the performance of a custom GPT in our study.

## 4. Data and analysis

All the data for the trial is from real classroom dialogue collected in the project that generated the CDAS scheme. The human coding results are from the original CDAS data analysis, and the two coders were trained on the scheme. The inter-coder reliability (the agreement) between the coders is around 70%-75%. A test set of 1,386 turns from three lessons was randomly selected as a testing data set, including one English lesson and two math lessons. The examples used for AI coder training and testing were selected from other datasets within the same project—the 144 lessons that comprise the CDAS project.

To evaluate the performance of the AI dialogic coder, we used a confusion matrix, a standard tool in machine learning that compares a model's predicted classifications with the actual ground truth labels (Visa et al., 2011). It provides a visual summary of classification outcomes by displaying the number of true positives (TP), true negatives (TN), false positives (FP), and false negatives (FN), enabling a detailed assessment of accuracy and error patterns. In this study, the outcomes were defined according to the agreement between the AI's coding and the human coders' results: true positives occurred when both the AI coder and human coders agreed that a turn should be coded as X; false negatives were cases where the human coders applied code X but the AI coder did not; false positives indicated the AI coder applied code X while human coders did not; and true negatives occurred when both the AI coder and human coders agreed that code X should not be applied.

To quantify the AI coder's performance, we calculated four key metrics. Precision measures the proportion of correctly predicted (the correctness here means the AI coder managed to reach an agreement with the human coders'

$$\text{Precision} = \frac{TP}{TP + FP}$$

results) positive instances out of all predicted positives:

Recall (or sensitivity) reflects the proportion of actual positives that were correctly identified:

$$\text{Recall} = \frac{TP}{TP + FN}$$

Accuracy indicates the overall proportion of correct predictions:

$$\text{Accuracy} = \frac{TP + TN}{TP + TN + FP + FN}$$

Finally, the F1 score, the harmonic mean of precision and recall, provides a balanced measure that accounts for both false positives and false negatives:

$$\text{F1 Score} = \frac{2 \times TP}{2 \times TP + FP + FN}$$

These metrics provide a comprehensive view of the AI coder's coding performance in relation to human judgments. However, in the absence of a definitive ground truth, we cannot assume that the human coders' annotations are inherently correct. Therefore, the reported precision and accuracy reflect the degree of alignment between the AI coder's outputs and the human coders' decisions, rather than serving as absolute indicators of the AI coder's correctness or validity.

## 5. Results

**RQ1) How does a MyGPT with basic configurations (instructions) perform in analysing classroom dialogue using an adapted version of SEDA (CDAS) for human coders?**

For the baseline test, the materials entered into the instruction box were taken from the non-justified version used in human coders' training sessions, including the explanations and keywords for the coding categories. An example is shown below:

" *Reasoning (RE): Provides an explanation or justification of own or another's contribution. Includes drawing on evidence (e.g. identifying language from a text/poem that illustrates something), drawing analogies (and giving reasons for them), making distinctions, breaking down or categorising ideas. It can include speculating, hypothesising, imagining and predicting, so long as grounds are provided. Keywords include 'because', 'if...then', 'so', 'therefore', 'not...unless', 'would', 'could', might'.*" (CDAS Coding Manual, 2019)

Other important rules added in the instruction include: "As CDAS coder, my primary function is to support deductive qualitative coding of classroom dialogues using the Cambridge Dialogue Analysis Scheme (CDAS)", "Coding should be based on the flow of meanings across turns, not partially on each sentence or each turn," and "Only code within these 13 codes. Each turn can be assigned more than one code.".

As shown in Table 2, the AI coder's overall performance in the baseline test remained low, with most precision values varying from 6.7% to 38.2%. The most successfully identified codes are RE (Reasoning), IRE (Invite Reasoning), and OI (Other Invitation). One possible explanation is that the first two of these codes were often marked by clear discourse indicators, such as 'why' and 'because'. Each code was evaluated independently, which contributes to the high accuracy rate, as correctly excluding a code is also counted as a success. Nevertheless, it is clear that the AI coder could not be deployed effectively without further configuration.

**Table 2**

*Confusion metrics of the AI coder with the baseline setting*

| Categories | Precision | Recall | Accuracy | F1 Score |
| --- | --- | --- | --- | --- |
| ELI | 16.3% | 20.0% | 90.1% | 18.0% |
| EL | 38.2% | 28.4% | 73.5% | 32.6% |
| IRE | 36.0% | 47.2% | 89.5% | 40.8% |
| RE | 67.2% | 33.1% | 85.5% | 44.3% |
| CI | - | - | 94.3% | - |
| SC | 6.7% | 25.0% | 98.8% | 10.5% |
| RC | - | - | 99.5% | - |
| A | 24.3% | 16.7% | 75.5% | 19.8% |
| Q | 9.0% | 17.5% | 92.5% | 11.9% |
| RB | - | - | 97.0% | - |
| RW | 7.7% | 12.5% | 98.0% | 5.9% |
| OI | 47.8% | 79.8% | 90.2% | 59.8% |
| Uncoded | 71.7% | 37.1% | 74.1% | 48.9% |

**RQ2) Does input size matter in MyGPT? How does its performance change with different small sizes of examples?**

The performance of the custom GPT model varies depending on the size and structure of the examples provided. As the number of training examples increases (from 12 to 120 and then to 500), the precision of the dialogic codes generally shows an upward trend. However, the most notable improvement occurs between 12 and 120 examples, with relatively modest gains beyond that point. This plateau may be caused by the model's token limit, as once the input reaches a certain length, the model is likely to condense and synthesise information rather than process each example in full detail. As a result, simply increasing the quantity of examples beyond the optimal threshold may yield diminishing returns.

While increasing the number of training examples tends to improve overall performance, the quality and context of the examples also play an important role. As shown in Table 3, using a diverse range of contextual dialogue samples, rather than isolated single turns or samples drawn from a single lesson, results in better precision for most dialogic codes. For instance, in the second round of training, contextual examples drawn from 120 diverse dialogue turns outperformed single-turn samples across several key codes such as ELI, EL, RE, and RB. This suggests that contextual depth helps the model generalize dialogic patterns more effectively, perhaps due to better representation of discourse flow and interactional cues.

However, certain dialogic codes—particularly those that may be more structurally consistent or formulaic—still benefit from more focused, single-turn examples. For example, code A (Agreement) achieved the highest precision (50%) in the second round with single-turn samples, compared to 34.5% with contextual samples and 33% in the larger third round. Similarly, codes such as Q (Query) and RW (Reference to wider context) also performed comparably with minimal examples, indicating that some behaviours may be more easily learnable without extensive context.

That said, even the third round, which used 500 training examples, did not yield uniformly better performance across all codes—some declined, such as OI (Other Invitation) and Uncoded, suggesting diminishing returns or overfitting in certain areas. These findings support the idea that while data size does matter, the structure, diversity, and dialogic richness of the examples are equally, if not more, important in optimising model performance.

**Table 3**

*Precision Across Rounds Using Varying Example Sizes for Training (Details of each condition are provided in Section 3.1)*

| Code | Condition 1 | Condition 2 | Condition 3 | Condition 4 |
|---|---|---|---|---|
| ELI | 16.3% | 12.70% | 21.3% | 14.7% |
| EL | 38.2% | 35.78% | 43.5% | 54.1% |
| IRE | 36% | 31.94% | 37.7% | 43.9% |
| RE | 67.2% | 49.2% | 51.5% | 70.5% |
| CI | - | - | - | - |
| SC | 6.7% | 5.9% | 0.0% | 0.0% |
| RC | - | - | - | - |
| A | 24.3% | 50.0% | 34.5% | 33.0% |
| Q | 9.0% | 7.0% | 6.7% | 12.2% |
| RB | - | 12.0% | 33.3% | 25.0% |
| RW | 7.7% | 7.7% | 50.0% | 100.0% |
| OI | 47.8% | 49.0% | 48.2% | 42.9% |
| Uncoded | 71.7% | 63.8% | 65.5% | 58.6% |

**RQ3) What are the potential strategies for building an effective MyGPT model for coding dialogues?**

After several rounds of trials following the conjectures summarised from the literature review, an AI coding assistant with an overall 64.9% turn-level agreement (counting both exact and partial agreement) was developed, which is approaching the generally desired level of 70% inter-coder reliability between human coders. The performance of the AI coder is presented in Table 4. This has the potential to be optimised further with the application of prompt engineering. Certain codes, such as A and Q, are frequently embedded in multi-utterance turns and are easy to ignore, requesting further adjustment using recoding with strategised prompts. An example of the final results for coding multi-utterance turns is shown below (Figure 1).

**Table 4**

*Final Outcomes from Implementing All Alternative Strategies*

| Categories | Precision | Recall | Accuracy | F1 Score |
|---|---|---|---|---|
| ELI | 51.1% | 32.1% | 90.1% | 39.5% |
| EL | 61.0% | 56.6% | 80.5% | 58.7% |
| IRE | 80.8% | 51.5% | 92.9% | 62.9% |
| RE | 46.9% | 73.1% | 88.3% | 56.5% |
| CI | - | - | - | - |
| SC | 0.0% | 0.0% | 99.4% | 0.0% |
| RC | - | - | - | - |
| A | 30.5% | 78.4% | 85.4% | 44.0% |
| Q | 12.8% | 55.5% | 97.2% | 20.8% |
| RB | 50.0% | 48.3% | 97.9% | 49.1% |
| RW | 33.3% | 21.4% | 98.8% | 26.1% |
| OI | 78.1% | 60.1% | 87.9% | 67.9% |

**Figure 1**

*Screenshot of MyGPT AI coder's Coding Outcome Example.*

**Turn 241 – Teacher**

**Transcript:**

*"Well done, you're absolutely right Maya, take away the 20%, the 92, and so we can do a column subtraction. £4.60 takeaway 92p, (( as column subtraction )) off you go. (( Turns to Marley who is standing waiting with a question )) Marley?"*

**Analysis:**

This is a **multi-utterance turn** containing:

1. **Agreement**: "You're absolutely right, Maya." → Affirming student input.
2. **Elaboration**: Adds factual info: "take away the 20%, the 92..."
3. **Procedural instruction**: "off you go."
4. **Shift to new student with a question**: "Marley?"

**Codes: A, EL, OI**

- **A** – Explicit agreement with Maya's (unheard) answer.
- **EL** – Adds factual elaboration with specifics of the subtraction method.
- **OI** – "Marley?" functions as an **invitation to contribute**, not linked to prior conceptual input. Since it's open and not specifically reasoning/elaboration-linked, it falls under **Other Invitation**.

To optimise the performance of the MyGPTs-based AI coder in coding dialogic interactions, we developed a set of targeted strategies involving prompt engineering, cognitive load management, and iterative refinements based on error analysis. These strategies (described below) are tailored to the specific challenges of rule-based annotation tasks, where precision, interpretability, and consistency are essential.

**Strategy 1: Segmented instruction**

After assigning the role: "You are a coding assistant specialising in applying the Cambridge Dialogue Analysis Scheme (CDAS) to analyse classroom dialogue interactions. Your task is to automatically assign the most relevant CDAS codes to classroom dialogues using a structured decision-making process. The coding scheme is based on the intention of the speaker, so interpreting the utterances in the context of the dialogue is critical," the instruction setting was segmented into modules, including:

***Module 1, Definition of codes***

Example: *Elaboration Invitation (ELI): Invites building on, elaborating, or clarifying prior input(s). Includes critique, compare, or evaluation prompts. ELI must follow from a previous utterance; if no link exists, code as OI. Excludes procedural or calculation questions (use OI).*

*- Example: "Can you say more about what the poet uses?" → ELI*

***Module 2, Decision tree***

Example: *Step 1: Verify the Learning Goal Before Coding*

*• If relevant, go to step 2 and process coding*

*• If utterance is not relevant to the learning goal → Uncoded*

*- Examples: Greetings, administrative actions, off-topic conversation.*

*- Example: *"Good afternoon."* → **Uncoded***

*- Example: *"Put your books away and sit down."* → **Uncoded** (procedural but unrelated to learning).*

***Module 3, Rules for justification***

Example: *Invitational codes (**ELI, IRE, CI, OI**) exclude rhetorical/non-verbal prompts.*

***Module 4, Stability control***

Example: *If a previous utterance was coded differently, prompt:*

*"A similar response was coded as [X]. Does this classification align?"*

**Figure 2**

*Structure of the customised GPT's Instruction Configuration*

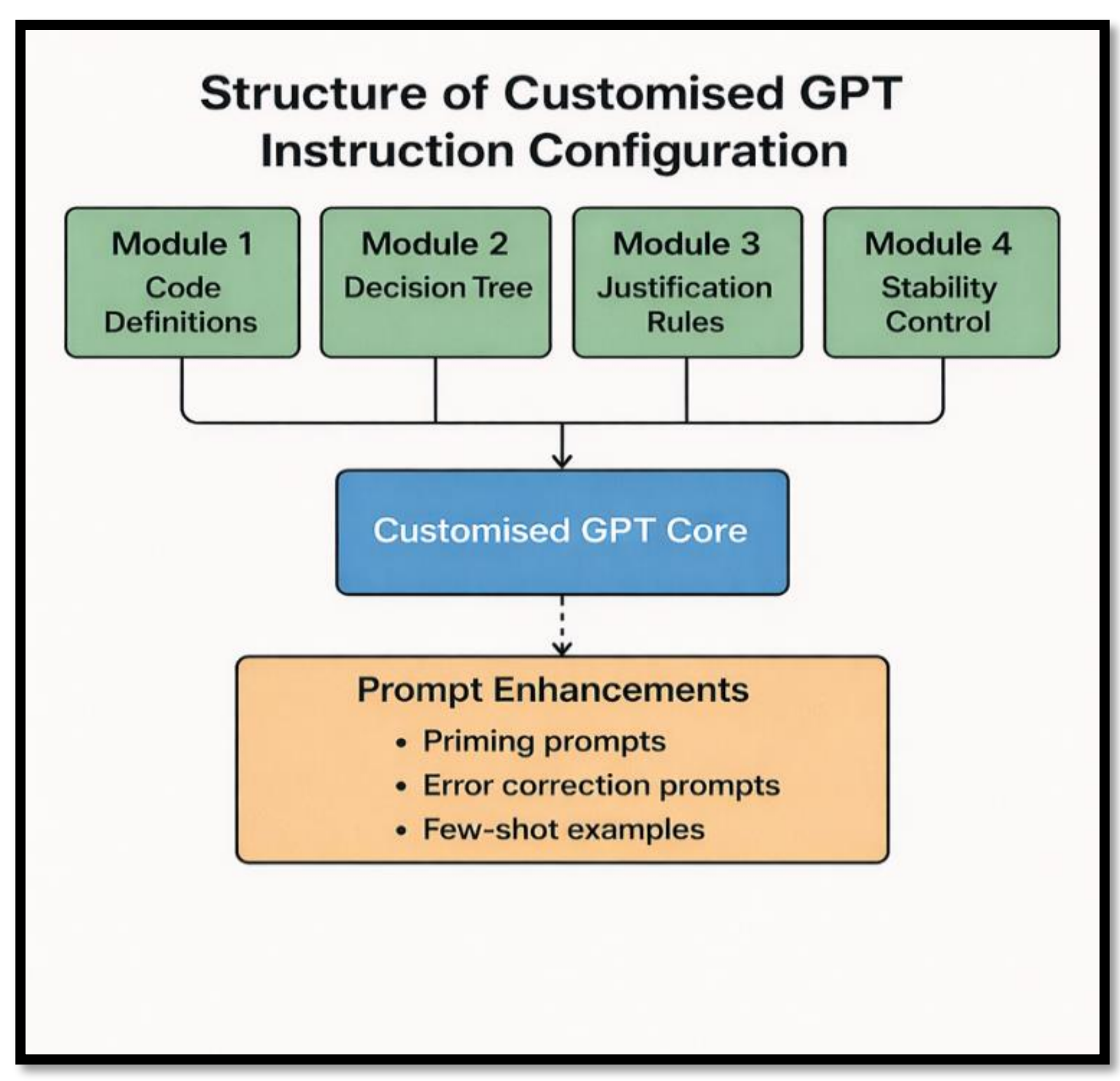

**Strategy 2: Be Aware of the Token Limit**

Due to GPT's inherent token limitations, the model can only process and retain a finite amount of information. Beyond a certain threshold, it begins to condense or abstract input data, which can lead to misinterpretations or loss of fidelity in decision-making. Thus, example input should be concise and carefully selected to maximise instructional clarity before hitting this ceiling.

**Strategy 3: GPT-Friendly Instructions Using Programmatic Thinking**

Instructions were crafted to mirror logic structures familiar to machine learning and programming, such as conditional statements and rule-based trees. By aligning instructions with machine-executable logic (e.g., "if-then" conditions), the model processes them more reliably, mimicking the structure of algorithmic reasoning.

**Strategy 4: Reducing Cognitive Load via Decision Trees**

Complex classification tasks involving overlapping or multi-layered coding rules tend to confuse the model. To mitigate this, we reduced cognitive load by organising instructions into hierarchical decision trees, allowing the model to follow a clear, step-by-step reasoning path.

**Strategy 5: Prioritisation and Instruction Order**

Earlier instructions appeared to have greater influence. Therefore, the most important rules and decision points are placed at the start of the prompt. Instructions are also ranked by priority to guide the model's decision-making process in ambiguous situations.

### Strategy 6: Interactive Rule Verification

To ensure the model's alignment with human coding logic, we used interactive sessions to test its understanding of the rules. For example, prompts such as "explain your understanding of coding multi-utterance dialogue" were used, and comments to confirm or adjust the understanding were provided. This "interactive building mode" allowed for early correction of misinterpretations before full-scale application.

### Strategy 7: Instruction-Embedded Core Examples

Essential examples were embedded directly into the instruction set to anchor the model's decision boundaries. We ensured diversity across types (e.g., clear, ambiguous, borderline cases) to reflect real-world complexity.

### Strategy 8: Iterative Feedback for Error Patterns

Frequent error categories (e.g., mislabelling of codes like A vs. EL, or misclassifying multi-utterance turns) were addressed through repeated feedback cycles. This iterative configuration via human-in-the-loop correction enabled the model to refine its understanding over time.

### Strategy 9: Instruction Weight > Example Quantity

In our observations, clear and logically structured instructions consistently outperformed large quantities of examples. While examples offer context, they cannot compensate for vague or overly complex instruction sets.

### Strategy 10: Influence of Previous Input

GPT's memory of recent inputs can bias subsequent performance. Inputs from prior conversations or prompts may implicitly overwrite or distort the current model behaviour. This highlights the importance of resetting sessions when needed.

**Strategy 11: Avoiding Overloaded Prompts and Thread Drift**

Long threads or excessive example sequences (>20 coding turns) introduce risks of drift, where GPT's outputs begin to deteriorate in coherence or consistency. Moreover, without feedback in these long threads, the model may overlearn from its own errors, reinforcing incorrect labelling patterns. To prevent this, we recommend working in smaller coding batches and restarting sessions periodically.

**Strategy 12: Self-Monitoring and Historical Comparison**

To improve consistency, the model can be prompted to check its own outputs using meta-cognitive prompts like *"Coding confirmed using decision tree."* It can also perform historical comparisons (e.g., "Previous responses matched as RE") to maintain intra-session consistency.

**Strategy 13: Optimal Number of Examples:**

- Core Examples: Include 10-15 examples, ensuring at least one for each code in CDAS. This covers the basic usage and key distinctions of each code. Considering the token limit, we recommend incorporating more examples within the feedback loops rather than providing a large number of examples during the initial tuning phase.
- Ambiguous Cases: Add 5-10 examples focusing on borderline or challenging scenarios (e.g., distinguishing ELI from OI or EL from RE).

- Multi-Utterance Turns: Include 5 examples to show how to handle turns with multiple utterances or mixed codes.
- Edge Cases: Add 5 examples for uncoded or context-specific cases, such as non-dialogic remarks, unfinished statements, or procedural turns.

**6. Discussion**

The present study explored diverse strategies for developing a MyGPT— an accessible and lightweight large language model—to automatically code classroom dialogues. The study employed an established coding framework applied to dialogue data from UK primary schools across three subjects (mathematics, science, and English), which had previously been annotated by trained human coders. Four key insights emerged from the findings.

First, providing the custom GPT with the original coding manual alone might not suffice. Under this simple approach, only one category (Reasoning) – characterised by clear discourse cues – approached acceptable accuracy (70%), excluding the "Uncoded" category. All other codes fell below a usable threshold, suggesting that a purely deductive, manual-based method designed for human coders is inadequate for achieving high-accuracy dialogue coding.

Second, enlarging the pool of worked examples can improve performance, but only up to a certain point. Increasing the number of human-annotated turns from 120 to 500 heightened accuracy in a few categories (i.e., EL, RE, Q, RW) yet yielded no enhancement – or even a decline in performance for the remainder. This decline is likely attributed to the MyGPT's token limits, which cap the number of tokens processed in a single prompt. When an excessive number of annotated turns are supplied, the AI coder may exhibit reduced coherence due to its cognitive overload (Yue et al., 2023). The use of a larger-window version might soften – but not remove – this ceiling.

Third, supplying contextualised excerpts (consecutive turns from the same exchange) produced higher accuracy for several key codes (e.g., ELI, EI, RE, and RB), compared with isolated single turns. What should not be overlooked, nonetheless, is that two codes (Q, A) were identified more accurately from single-turn examples. This variability indicates that contextualisation may be beneficial for dialogic moves that rely largely on inter-turn relationships, but its advantages can be mediated by the semantic granularity of each code – that is, how specific and context-dependent the meaning of each code is.

Last but not least, the most notable enhancement resulted from 1) structuring the instructions to mitigate cognitive load (e.g., via decision trees with explicit IF-THEN conditions segmented instructions), 2) iteratively refining the instructions based on error analysis, and 3) using a 'dialogue coding-only' reminder to prevent the AI coder from defaulting to Chat GPT's open-ended conversational mode. These results align with prior research on LLMs' cognitive scaffolding – structured prompt design that supports GPT's reasoning more like a trained human coder (Yao et al., 2023). They further suggest that developing an effective configured AI system can be a recursive – partly inductive – process: errors should be identified, prompts continually refined until its performance reaches a satisfactory level, and initial coding rules stressed (Long et al., 2024).

## 7. Limitations

This study's findings should be interpreted with caution due to several contextual and technical limitations. It used transcripts from UK primary classrooms in core subjects, where dialogic practices and curricular norms may not be typical of other educational stages, subjects, or cultural settings. The work was conducted using a MyGPT, which has specific technical constraints that influenced the prompt strategies; these may not directly transfer to other LLM platforms with different architectures or token limits. Lastly, the model's internal decision-making process remains opaque, restricting understanding of how prompt refinements affect its reasoning. These limitations suggest that the reported strategies may not be universally applicable and underscore the need for further research to assess their external validity and interpretability.

## 8. Implications and contributions

This study reveals that a low-cost, lightweight custom GPT can function as a reliable tool for dialogue coding – if it is purposefully engineered rather than merely provided with a manual intended for human coders. Effective automation of coding via MyGPTs requires prompts tailored to such AI systems' technical constraints (e.g., token limits, cognitive load, and their tendency to revert to open-ended conversation). Strategies including structured prompts, context-sensitive examples, and iterative refinement proved helpful in attaining usable performance, without depending on expensive infrastructure, custom LLM development or extensive configurations. Importantly, the findings suggest that acceptable performance can be achieved even when datasets annotated by human coders are limited – a common constraint in real-world research and educational settings. By disclosing lower-barrier and replicable approaches, the study highlights the potential to expand access to automated dialogue analysis, empowering researchers and teachers without expertise in LLMs to create usable tools for dialogue coding.